\documentclass[wcp]{jmlr}

\usepackage{longtable}
\usepackage{booktabs}
\usepackage{hyperref}
\usepackage{amsmath}
\usepackage{amssymb}
\usepackage[hyphenbreaks]{breakurl}
\usepackage[misc]{ifsym}

\newcommand{\ve}[1]{\mathbf{#1}} 

\makeatletter
\let\Ginclude@graphics\@org@Ginclude@graphics
\makeatother
\jmlrvolume{}

\title[AIIR-MIX]{AIIR-MIX: Multi-Agent Reinforcement Learning Meets Attention Individual Intrinsic Reward Mixing Network}

\author{\Name{Wei Li$^{(\textrm{\Letter})}$} \Email{li-wei@seu.edu.cn}\\
\Name{Weiyan Liu} \Email{liuweiyan@seu.edu.cn}\\
\Name{Shitong Shao} \Email{shaoshitong@seu.edu.cn}\\
\Name{Shiyi Huang} \Email{huangshiyi@seu.edu.cn}\\
\addr School of Instrument Science and Engieering, Southeast University, Nanjing, Jiangsu 210096, China
}

\begin{document}
\begin{sloppypar}
\maketitle

\begin{abstract}
Deducing the contribution of each agent and assigning the corresponding reward to them is a crucial problem in cooperative Multi-Agent Reinforcement Learning (MARL). Previous studies try to resolve the issue through designing an intrinsic reward function, but the intrinsic reward is simply combined with the environment reward by summation in these studies, which makes the performance of their MARL framework unsatisfactory. We propose a novel method named Attention Individual Intrinsic Reward Mixing Network (AIIR-MIX) in MARL, and the contributions of AIIR-MIX are listed as follows: \textbf{(a)} we construct a novel intrinsic reward network based on the attention mechanism to make teamwork more effective. \textbf{(b)} we propose a Mixing network that is able to combine intrinsic and extrinsic rewards non-linearly and dynamically in response to changing conditions of the environment. We compare AIIR-MIX with many State-Of-The-Art (SOTA) MARL methods on battle games in StarCraft II. And the results demonstrate that AIIR-MIX performs admirably and can defeat the current advanced methods on average test win rate. To validate the effectiveness of AIIR-MIX, we conduct additional ablation studies. The results show that AIIR-MIX can dynamically assign each agent a real-time intrinsic reward in accordance with their actual contribution.
\end{abstract}

\section{Introduction}
Deep Reinforcement Learning (DRL) is an crucial branch of machine learning. It utilizes neural networks to approximate the optimal action decision or value function of the agent, realizing the generalization of the representation ability. By reason of the powerful fitting ability of deep learning, DRL can be employed as an effective way to tackle agent decision-making problems in complicated environments, such as group decision-making~\cite{Nguyen2020deep}, speech recognition~\cite{mousavi2016deep, shen2019reinforcement}, autonomous driving~\cite{shen2019reinforcement}, natural language processing~\cite{young2018recent}, and intelligent control~\cite{carlucho2020adaptive}.

A multi-agent environment, where multiple agents are present for interaction and learning, is an emerging hot topic in recent years. Unfortunately, single-agent reinforcement learning is not very effective when used in a multi-agent environment, because the joint action space of the agent resulting from fully centralized learning is too large to learn the optimal policy. Therefore, Multi-Agent Reinforcement Learning (MARL) is an extension of deep reinforcement learning from single-agent to multi-agent, and has a list of methods to solve this problem. Centralized Training and Decentralized Execution (CTDE)~\cite{foerster2016learning, wang*2020influence-based}, an effective and widely applied method in this list. In this paradigm, the central controller manages all the agents' observations, actions, and rewards during training. And the central controller and its value networks are not utilized during execution. Due to the above features and superiority, CTDE has become a widely applied paradigm in MARL, such as COMA~\cite{foerster2018counterfactual}, VDN~\cite{sunehag2017value}, QMIX~\cite{rashid2018qmix} and QTRAN~\cite{son2019qtran}. In our research, we will also follow the CTDE paradigm in our proposed method.

In MARL, the reward function is extremely significant. However, it is difficult to go about formalizing all situations as reward functions in some real-world tasks. For exploring new environments in reinforcement learning, often only sparse rewards can be set. And yet, in practical application scenarios, sparse rewards still face problems such as inefficient samples and difficulty in exploration. This problem is more evident in MARL. Because of the inherent challenges of MARL, such as unstable environment and dimensional catastrophe, extending MARL to sparse reward settings will further increase the difficulty of policy learning. A good method for solving the incentive issue in a multi-agent environment is to assign extrinsic rewards and construct an intrinsic reward for each agent. In recent years, many researchers have started to focus on these directions. LIIR~\cite{du2019liir} learns an intrinsic reward function for each agent and continuously updates it to maximize the expected accumulated team reward from the environment. GIIR~\cite{wu2021generating} solves the lazy agent problem by using an intrinsic reward encoder to generate a separate intrinsic reward for each agent. OpenAI~\cite{berner2019dota} uses artificially set intermediate rewards to accelerate learning. FTW~\cite{jaderberg2019human} learns the agent's intrinsic reward through two layers of optimization. However, both approaches overlook the following points: \textbf{(a)} building dependencies (i.e., attention mechanisms) between agents can induce more precise rewards; \textbf{(b)} integrating intrinsic and extrinsic rewards to facilitate policy learning better.

In this paper, we propose Attention Individual Intrinsic Reward Mixing Network (AIIR-MIX) method to fill this gap. AIIR-MIX includes the generation of precise intrinsic reward network (AIIR) and a non-linear Mixing network (MIX) to combine the intrinsic and extrinsic rewards. In terms of generating intrinsic reward, we propose an intrinsic reward network based on the attention mechanism. We assume that each agent has a separate intrinsic reward. The agents' observations and actions are extremely similar when they perform teamwork. Based on the above, the contribution of each agent in teamwork is calculated from each agent's observation and action using the attention mechanism, and a more accurate intrinsic reward is generated for each agent. The intrinsic reward network is updated to maximize the standard cumulative discounted extrinsic rewards from the environment. In terms of combining intrinsic and extrinsic rewards, we propose a Mixing network that allows intrinsic and extrinsic rewards to be combined in a non-linear manner. The extrinsic reward is fed to the hyper network to generate the weights of the Mixing network. The Mixing network combines weights and intrinsic rewards to output global rewards for each agent. In addition, we apply an intrinsic reward function to the Actor-Critic algorithm, where each agent's individual policy is updated under the direction of the corresponding proxy critic. Benefitting from these improvements, AIIR-MIX generates a more appropriate global reward and reduces artificial intervention in reward function design.

We evaluate the AIIR-MIX method on StarCraft II micromanagement benchmark ~\cite{samvelyan2019starcraft}. The experimental results show that the AIIR-MIX performs better than the mainstream algorithms such as LIIR and QMIX in both homogeneous and heterogeneous maps. We conduct ablation experiments and demonstrate that both AIIR and MIX perform better than the baseline algorithm when they are used individually. Moreover, we visualize the training process and show the dynamic change process of attention weights and intrinsic reward as time advances in the complete trajectory. The results demonstrate the effectiveness and importance of the intrinsic reward based on the attention mechanism.

\section{Related Work}

\subsection{Sparse Reward}
In many Reinforcement Learning (RL) tasks, the reward from the environment is sparse. Therefore, it may take agents too many steps to reach the state with a positive reward. This situation causes many problems such as low efficiency and exploration difficulty.

Researchers have proposed an intrinsic motivation approach to address the sparse reward problem, in which an intrinsic reward function will be designed to generate intrinsic rewards for agents to promote learning efficiency.~\cite{pathak2017curiosity} proposed using prediction errors in pixel space as curiosity rewards to drive exploration in a self-supervised manner.~\cite{strehl2008analysis} recorded counts of accessed state-action pairs in table form and converted the counts into an intrinsic reward, which was additionally added to the reward from environmental feedback.~\cite{song2018multi} extended generative adversarial imitation learning to the multi-agent field, but the need for expert demonstrations limited its generality.~\cite{hao2019independent} combined generative adversarial imitation learning and self-limitation learning and applied them to multi-agent systems to facilitate multi-agent cooperation, but still did not fundamentally reduce the difficulty of training when the number of agents was large.

In order to design an intrinsic reward function that can solve the sparse reward issue more efficiently, the attention mechanism, a method of establishing dependencies between agents, can be utilized by us to solve the sparse reward issue in MARL.

\subsection{Attention mechanism}
The attention mechanism, a method capable of automatically selecting significant information, is widely utilized in computer vision, natural language processing, and reinforcement learning. In recent years, the attention mechanism has been introduced into MARL to facilitate the learning effectiveness of agents to some extent. For example, Multi-Actor-Attention-Critic (MAAC)~\cite{iqbal2019actor} applies an attention mechanism to model centralized Critic networks. The Attention communication (ATOC)~\cite{jiang2018learning} model proposes a bi-directional LSTM communication channel with an attention layer, whose attention mechanism allows each agent to focus on messages from other agents according to their state-related importance.

Inspired by these work, we propose an intrinsic reward generation network based on an attention mechanism. Unlike existing methods, this paper utilizes an attention mechanism in generating intrinsic rewards for each agent. It adaptively processes historical information from other agents, focuses on each agent's contribution in teamwork, and generates more precise intrinsic rewards. At the same time, the intrinsic reward network is combined with the Actor-Critic architecture to improve its performance.

\section{Background}
In generally, a fully cooperative multi-agent problem can be described as a decentralized partially observable Markov decision process (Dec-POMDP)~\cite{oliehoek2016concise} consisting of a tuple $G=<\mathcal{N},S,U,P,Z,O,r,\gamma ,\rho _{0}>$. $\mathcal{N}=\left\{1,2,\cdots ,n\right\}$ denote the set of $n$ agents and $i\in \mathcal{N}$. $s\in S$ is the state of the environment which includes global information for all agents. $U$ is the set of actions. $Z$ is the agent's observation set, each agent gets its own observation $o_{i}\in Z$ according to the observation function $O(s,i):S\times \mathcal{N}\rightarrow Z$. At each timestep, each agent $i\in \mathcal{N}\equiv \left\{1,\cdots ,n \right\}$ chooses an action $u_{i}$ through the parameterized policy network $\pi_{i}\left(o_{i}\right) $ according to the current observation $o_{i}$, forming a joint action $u_i\in U\equiv U^{n}$ and leading to next state $s'$ according to the transition function $P(s'|s,u):S\times U\times S\rightarrow \left [ 0,1 \right ]$. $\boldsymbol{\pi }= \left\{\pi _{1},\pi _{2},\cdots,\pi _{n} \right\} $ denotes the joint policy consists of the policy of each agent. In order to distinguish different rewards, we denote the team reward from the environment as extrinsic reward $\ve{r}^{\textrm{ex}}$. The intrinsic reward set that will be learned as $\ve{r}_t^{\textrm{in}}={\left\{r^{\textrm{in}}_{i}\right\}}_{i=1}^{n}$, where $t$ is the index of the timestep. $\ve{r}^{\textrm{ex}}(s,u_i):S\times U\rightarrow \mathbb{R}$ is the team reward for each agent $i$ from environment. $\rho_{0}:S\to \mathbb{R}$ is the distribution of the initial state $s_{0}$. In a fully cooperative multi-agent problem, each agent receives the same $\ve{r}^{ex}$ to promote cooperative behavior.

The learning objective of the cooperative multi-agent problem is that $n$ agents learn a policy network $\pi_{i} $ parameterized by $\theta_{i} $ to maximize the global cumulative discounted reward set $\ve{r}_t^{\textrm{total}}$. That is, when $\ve{r}_t^{\textrm{total}}$ is the largest, the optimal joint policy of all agents is obtained $\boldsymbol{\pi^*}= \mathop{\textrm{argmax}}\limits_{\pi}\mathbb{E}_{s_0,u_0,...,s_n,u_n}\left[\sum_{t= 0}^{T}\gamma ^{t}\ve{r}^{\textrm{ex}}_{t}\right]$, where $\gamma\in\left[0,1\right)$ is a discount factor and $T$ is the maximum number of steps.

\subsection{Policy Gradient \label{3.1}}
The goal of reinforcement learning is to find an optimal behavior policy for the agent so as to obtain the maximum reward. The main characteristic of the policy gradient method is to model and optimize the policy directly. A policy is usually modeled as a function $\pi_{\theta } (a|s)$ parameterized by $\theta$. For each training iteration, the parameter $\theta$ is changed in the direction given by the gradient $\triangledown_{\theta} J(\theta) $ to find the optimal $\theta^*$, the gradient related to the parameter is expressed as:
\begin{equation}
\begin{aligned}
\triangledown_{\theta }J(\theta )= \mathbb{E}_{s\sim d^{\pi_\theta},u\sim \pi_\theta}\left [\mathbf{\Psi}(s,u)\triangledown _{\theta }\log \pi _{\theta }(u|s) \right ],
\end{aligned}
\end{equation}
where $d^{\pi_\theta}$ denotes a state transition following the policy, and $\mathbf{\Psi}(s,u)$ is the trajectory reward related to agent states and actions. The policy gradient algorithm is extended to multi-agent field, and each agent $i\in \left\{1,\cdots ,n \right\}$ has a policy function $\pi_{\theta _{i}} (u_{i}|o_{i})$. The multi-agent policy gradient can be expressed as:
\begin{equation}
\begin{aligned}
\triangledown_{\theta } J(\theta )= \mathbb{E}_{\pi_{\theta_i}\sim\boldsymbol{\pi_{\theta}}}\left [\sum_{i=1}^{n} \mathbf{\Psi}(s,\textbf{u})\triangledown _{\theta _{i}}\log \pi_{\theta _{i}}(u_{i}|o_{i}) \right ],
\end{aligned}
\end{equation}
where $u_{i}$ denotes the action of agent $i$, $o_{i}$ denotes the observation of agent $i$, $\textbf{u}$ and $\boldsymbol{\pi }$ denotes the joint action and policy of all agents, respectively. There are two training methods for calculating $\mathbf{\Psi}(s,\textbf{u})$. One is to train the policy network through the REINFORCE~\cite{williams1992simple}. The other is to introduce the policy gradient into the Actor-Critic framework, and actors are trained through the gradient of Critic. In this paper, we choose the second method in training. The advantage function, as a common way to measure the value of an agent action, is usually designed as $A_{\boldsymbol{\pi }}(s,\textbf{u})= Q_{\boldsymbol{\pi }}(s,\textbf{u})-V_{\boldsymbol{\pi }}(s)$, where $Q_{\boldsymbol{\pi }}(s,\textbf{u})$ is $r(s,\textbf{u})+\gamma V_{\boldsymbol{\pi }}(s')$ used to help calculate the advantage function.

\subsection{QMIX}
QMIX~\cite{rashid2018qmix} applies a mixing network that takes the outputs of all agent networks as input, mixes them monotonically, and adds global state information to the training process to improve algorithm performance. In this method, the weights and biases of the mixing network are generated by each hypernetwork based on the input state $s$. The generated weights ensure that the weights are non-negative by means of an absolute activation function. For each training iteration, the goal of QMIX is to minimize the loss function:
\begin{equation}
\begin{aligned}
\mathcal{L}(\theta )= \sum_{i=1}^{b}\left [ (y_{i}^{tot}-Q_{tot}(\boldsymbol{\tau} ,\ve{u},s;\theta ))^{2} \right ],
\end{aligned}
\end{equation}
where $y_{i}^{tot}= r+\gamma \textrm{max}_{u'}Q_{tot}(\boldsymbol{\tau},\ve{u}^\prime,s';\bar{\theta _{i}})$, $\bar{\theta_{i}}$ represent the parameters of the target networks, $b$ denotes the batch size of transitions sampled from the replay buffer, and $\boldsymbol{\tau}$ denotes the joint action-observation history. Inspired by this paradigm, we adopt a feed-forward neural network as a mixing network, combining $\ve{r}^{\textrm{in}}_{i}$ and $\ve{r}^{\textrm{ex}}$ into the global reward $\ve{r}^{\textrm{total}}_{i}$, thus improving the algorithm performance.

\subsection{Attention Mechanism}
For attention mechanism, the input is some vectors $\mathcal{V}\equiv\left\{\ve{v}_{1},\ve{v}_{2},\cdots,\ve{v}_{n}\right\}$. For each vector $\ve{v}_{i}$ in $\mathcal{V}$, according to Equation~\ref{eq:attention}, the output of attention mechanism is $\ve{\widetilde{v}}_i$.

\begin{equation}
\ve{\widetilde{v}}_{i}= \sum_{j=1}^{n}\frac{\textbf{\textrm{exp}}\left ( \sigma \left ( v_{i},v_{j} \right ) \right )v_{j}}{\sum_{k=1}^{n}\textbf{\textrm{exp}}\left ( \sigma \left ( v_{i},v_{j} \right ) \right )},
\label{eq:attention}
\end{equation}

where $\sigma\left(\cdot,\cdot\right)$ denotes a similarity metric function. AIIR-MIX devises an attention information processing mechanism to calculate the correlation of the historical information of the agents, so that it has the ability to adaptively identify and process meaningful information, and enhance team cooperation between agents.

\begin{figure}[t]
\begin{center}
\includegraphics[width=0.96\textwidth]{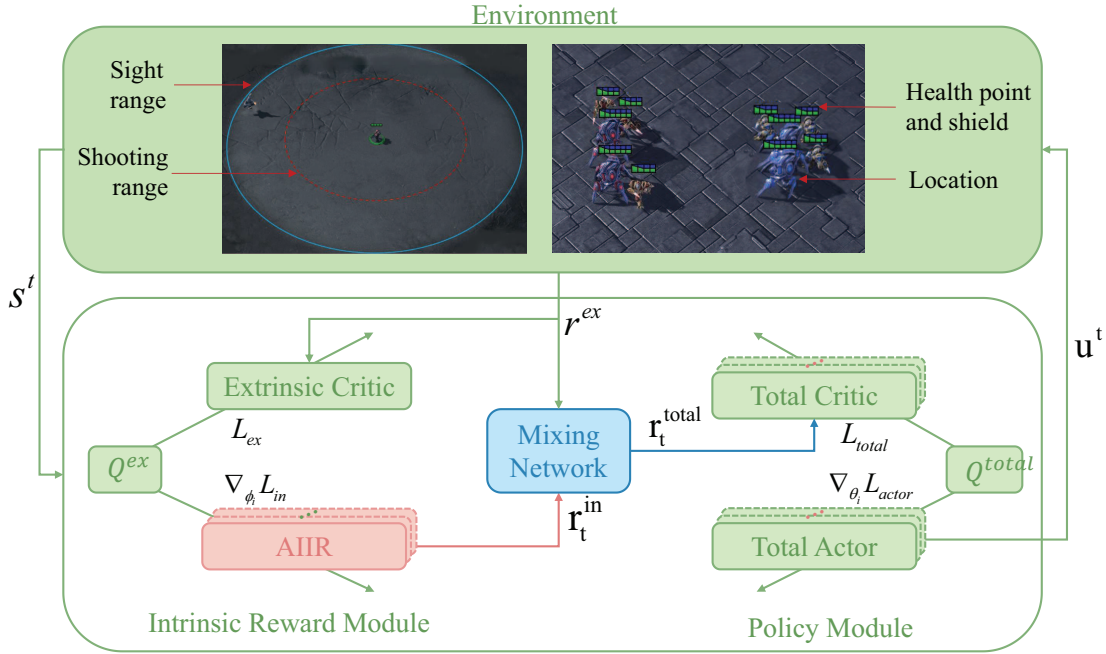}
\caption{Architecture of the overall AIIR-MIX framework. At timestep $t$, the total Actor obtains state information (e.g., health, location, shield, agent's sight range, and others) from the environment and generates actions. Then the $\ve{r}^{\textrm{ex}}$ is obtained from the environment and the intrinsic reward set $\ve{r}_{t}^{\textrm{in}}$ is generated by AIIR according to state information and actions. And Mixing network obtains the global reward set $\ve{r}_{t}^{\textrm{total}}$ by combining $\ve{r}_{t}^{\textrm{in}}$ and $\ve{r}^{\textrm{ex}}$ through non-linear operations.}\label{fig:1}
\vspace{-30pt}
\end{center}
\end{figure}

\section{Method}
In this Section, we introduce a new method called AIIR-MIX, which is based on the Actor-Critic framework. The main contribution of AIIR-MIX is to generate set $\ve{r}_t^{\textrm{in}}$ (w.r.t., the timestep is $t$) and combine it with $\ve{r}^{\textrm{ex}}$ to generate set $\ve{r}_t^{\textrm{total}}$ nonlinearly. We first present the overall AIIR-MIX framework in Fig.~\ref{fig:1}, and then we specifically show the relevant details about AIIR-MIX in Fig.~\ref{fig:2}.

As shown in Fig.~\ref{fig:1}, the total AIIR-MIX framework consists of five parts: extrinsic Critic network, total Critic network, total Actor network, AIIR, and Mixing network. AIIR and Mixing network make up a module for generating $\ve{r}_t^{\textrm{total}}$, which is the most important part of AIIR-MIX. In particular, the total Actor network is made up of $n$ Actor networks, each Actor network with parameter $\theta _{i}$ generates the policy of its corresponding agent. And we denote the current states $s_{i}$ as the input and the agent's policy $\pi_{\theta _{i}} (u_{i}|s_{i})$ as the output of each Actor network. Similarly, each Critic network in total Critic with parameter $\omega_{i}$ evaluates an agent's policy, which can update the Actor network end-to-end. Specifically, each Critic network updates the value function parameters $\omega_{i} $ depending on $Q(s_{i},u_{i};{\omega_{i}})$ and each Actor network updates the policy parameters $\theta_{i} $ for $\pi_{\theta _{i}}(u_{i}|s_{i}) $ in the direction suggested by its corresponding Critic network. Following the CTDE paradigm, the total Critic network is only utilized during training. Like~\cite{MADDPG,foerster2018counterfactual}, the total Critic network is updated by the Temporal-Difference error (TD-error) as well. But we utilize $\ve{r}^{\textrm{total}}_t$ generated from Mixing network by combining $\ve{r}^{\textrm{ex}}$ and $\ve{r}_t^{\textrm{in}}$ instead of $\ve{r}^{\textrm{ex}}$ to calculate TD-error. To better facilitate teamwork among the agents, we designed AIIR based on an attention mechanism to generate $\ve{r}_t^{\textrm{in}}$. Due to the powerful representational abilities of AIIR-MIX, it can update the total Critic network and the total Actor network better. 

Based on the Bellman equation, the loss function of a Critic network can be defined as:
\begin{equation}
\begin{aligned}
\mathcal{L}_{\textrm{total}}\left (\omega _{i}\right)= \frac{1}{n}\sum_{i=1}^{n}\left[{(y_{i}^{\textrm{total}}- Q(s_{i},u_{i};\omega _{i}))}^{2} \right],
\end{aligned}
\end{equation}
where $i\in \mathcal{N}\equiv \left\{1,\cdots ,n \right\}$ and $y_{i}^{\textrm{total}}=r_{i}^{\textrm{total}}+\gamma Q(s_{i}^{'},u_{i}^{'};\omega _{i}^{'})$. Of particular note is our definition of $r_{i}^{\textrm{total}}$ is the global reward of agent $i$, which is an element in $\ve{r}_{t}^{\textrm{tota}l}$. Each Actor network is updated by the policy gradient:
\begin{equation}
\begin{aligned}
\triangledown _{\theta_{i} }J\left(\theta _{i}\right)= E_{\pi _{\theta _{i}}}\left [ A_{\pi _{\theta _{i}}}\left(u_{i}|s_{i}\right)\triangledown_{\theta _{i}}\textrm{log}\pi_{\theta _{i}}\left(u_{i}|s_{i}\right) \right ],
\end{aligned}
\end{equation}
where $A_{\pi _{\theta _{i}}}\left(u_{i}|s_{i}\right)$ is the advantage function based on $A_{\boldsymbol{\pi }}(s,\textbf{u})$ (have been introduced in Section~\ref{3.1}) and $A_{\pi _{\theta _{i}}}\left(u_{i}|s_{i}\right)= r^{ex}\left(s_{i},u_{i}\right)+\gamma V\pi _{\theta _{i}}\left(s_{i}\right)-V\pi _{\theta _{i}}\left(s_{i}'\right)$. With a learning rate of $\alpha $ for the Actor network, the update of the parameter $\theta_{i}$ can be defined as:
 \begin{equation}
\begin{aligned}
\theta_{i,t+1}\leftarrow \theta _{i,t}+\alpha  A_{\pi _{\theta _{i}}}\left(u_{i}|s_{i}\right)\triangledown_{\theta _{i}}\textrm{log}\pi_{\theta _{i}}\left(u_{i}|s_{i}\right).
\end{aligned}
\end{equation}

By referring to Fig.~\ref{fig:1}, all agent share the extrinsic Critic network. The loss function of this extrinsic Critic network with parameter $\eta$ can be defined as:
\begin{equation}
\begin{aligned}
&\mathcal{L}_{ex}\left(\eta\right)= \left [ y^{ex}-Q(\ve{s},\ve{u};\eta )  \right ]^{2},\\
& \textbf{\textrm{where}} \ \ve{s}\equiv{\left\{s_i\right\}}_{i=1}^n, \ve{u}\equiv{\left\{u_i\right\}}_{i=1}^n, y^{ex} = \ve{r}^{ex}+\gamma Q(\ve{s'},\ve{u'};\eta ').
\end{aligned}
\end{equation}

Then, the generating procedure of $\ve{r}_t^{\textrm{in}}$ and $\ve{r}_t^{\textrm{total}}$ is described in detail. To promote cooperation amongst agents and meet the objective of maximizing the global reward for $\ve{r}_t^{\textrm{in}}$, we design an intrinsic reward generation framework based on an attention mechanism.

As shown in Fig.\ref{fig:2}\textcolor{blue}{c}, we first define a feature extractor, to process the state $s_{i}^{t}$ and the last action $u_{i}^{t}$ of the agent $i$ at timestep $t$, where the feature extractor is a sequence of consecutive ReLU-FC-ReLU-FC. Then, $\ve{v}^t_i$ appears both as an output of the feature extractor and as an input to the attention mechanism, representing the local attention embedding of the agent. With the help of the attention mechanism, we get the correlation of all agent pairs. Commonly, the correlation of agent pairs is obtained by a distance metric function, and it can be denoted in three forms as follows:
\begin{figure}[t]
\begin{center}
\includegraphics[width=0.96\textwidth]{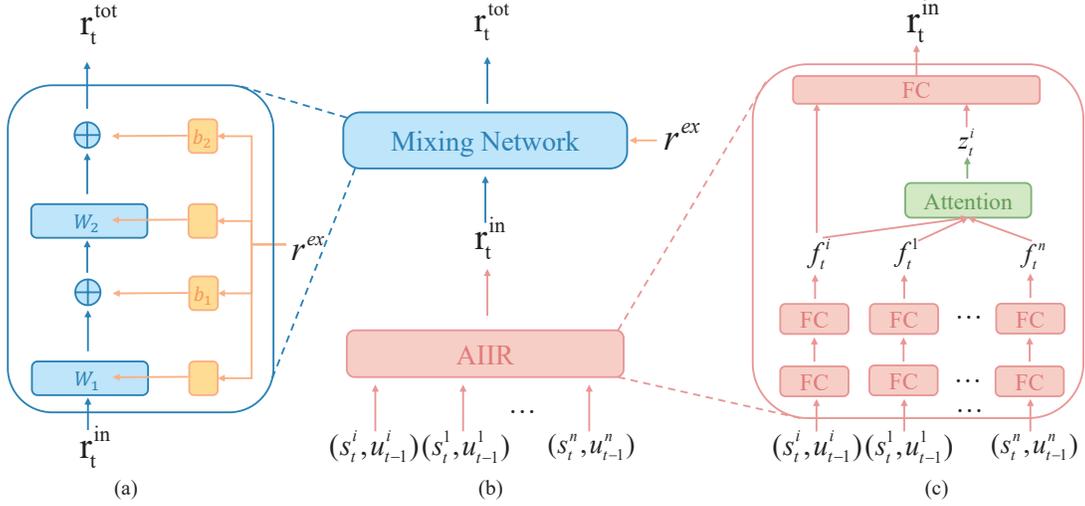}
\caption{\textbf{(a)} Mixing network structure. \textbf{(b)} The overall architecture for generating the global reward $\ve{r}_{t}^{\textrm{total}}$. \textbf{(c)} AIIR structure.}\label{fig:2}
\vspace{-30pt}
\end{center}
\end{figure}
\begin{equation}\small
\textrm{A}^{t}_{i,j}=\sigma\left(\ve{v}^t_{i},\ve{v}^t_{j}\right),\textbf{\textrm{where}}\ \sigma\left(\ve{a},\ve{b}\right)=\left\{
\begin{aligned}
&\ \ \! <\!\ve{a},\ve{b}\!> &,& \quad\ \quad \quad \quad \textrm{dot} ,& \\
&\frac{<\!\ve{a},\ve{b}\!>\  \ }{{\parallel\ve{a}\parallel}_2\!\cdot\!{\parallel\ve{b}\parallel}_2} &,& \quad\ \quad \quad\textrm{cosine},& \\
&\textrm{MLP}\left(\ve{a},\ve{b}\right) &,& \textrm{MLP\ network},&\\
\end{aligned}\right.
\end{equation}
since cosine similarity has the ability to normalize, we apply it to calculate the correlation between the attention embeddings of the agents.
In addition, we perform softmax function on the calculated correlations:
\begin{align}
\begin{split}
\mathbf{\widehat{A}}^{t}_{i,j}= \textrm{softmax}\left(\textrm{A}^{t}_{i,:}\right)= \frac{\textrm{exp}\left(\textrm{A}^{t}_{i,j}\right)}{\sum\limits_{k}\textrm{exp}\left(\textrm{A}^{t}_{i,k}\right)}.
\end{split}
\end{align}

The global attention embeddings $\ve{z}_{t}^{i}$ of agent $i$ is obtained by weighting and summing the attention embeddings of all agents according to the weighting coefficients:
\begin{equation}
\begin{aligned}
\ve{z}_i^{t}= \textrm{attn}\left(\mathbf{\widehat{A}}^{t}_{i,:},\ve{v}^t_i\right)= \sum_{j}\mathbf{\widehat{A}}^{t}_{i,j}\cdot \ve{v}^{t}_{i,j},
\end{aligned}
\end{equation}
where $\ve{z}_{i}^{t}$ contains the degree of similarity in the states and actions of the agents, learns the correlations between agents and promotes teamwork. After that, the local attention embedding $\ve{v}^{t}_{i}$ of agent $i$ and the global attention embedding $\ve{z}^{t}_{i}$ obtained through the attention mechanism are simultaneously provided as inputs to the fully connected layer. Finally, $r_{i}^{\textrm{in}}$ of agent $i$ and the set of all intrinsic rewards $\ve{r}_{t}^{\textrm{in}}$ are output:
\begin{equation}
\begin{aligned}
\ve{r}_t^{\textrm{in}}={\left\{r_i^{\textrm{in}}\right\}}_{i=1}^{n},\ \textbf{\textrm{where}}\ r_{i}^{\textrm{in}}= \textbf{\textrm{FC}}(\ve{z}^{t}_{i}).
\end{aligned}
\end{equation}

As shown in Fig.~\ref{fig:2}\textcolor{blue}{a}, $\ve{r}_t^{\textrm{in}}$ output by AIIR is non-linearly combined with $r^{\textrm{ex}}$ through the Mixing network to output $\ve{r}_{t}^{\textrm{total}}$. In previous studies, both LIIR~\cite{du2019liir} and GIIR~\cite{wu2021generating} combined $\ve{r}_t^{in}$ and $\ve{r}^{\textrm{ex}}$ by weighted summation:
\begin{equation}
\begin{aligned}
\ve{r}_{t}^{\textrm{total}}= \ve{r}^{\textrm{ex}}+ \lambda\ve{r}_{t}^{\textrm{in}}.
\end{aligned}
\end{equation}

In AIIR-MIX, we apply a non-linear approach to combine $\ve{r}_t^{\textrm{in}}$ and $r^{\textrm{ex}}$ to dynamically generate $\ve{r}_{t}^{\textrm{total}}$, so that the agent can obtain a more accurate reward at each timestep, thus facilitating the agent to select the optimal policy more readily. The weights $\{\mathcal{W}_{1},\mathcal{W}_{2}\}$ and biases $\{\ve{b}_1,\ve{b}_2\}$ of the Mixing network are generated by a separate hyper network, as shown in the left of Fig.~\ref{fig:2}\textcolor{blue}{a}. The $r^{\textrm{ex}}$ is utilized as input to the hyper network, which generates same weights $\{\mathcal{W}_{1},\mathcal{W}_{2}\}$ and biases $\{\ve{b}_1,\ve{b}_2\}$ through different linear layers and outputs them as weights and biases of the Mixing network, respectively. Combining the gradient descent method and the hyper network to update the Mixing network, allows for the dynamic adjustment of the weights of the Mixing network, thus enabling to combine $\ve{r}_t^{\textrm{in}}$ and $r^{\textrm{ex}}$:
\begin{equation}
\begin{aligned}
\ve{r}_{t}^{total}= {\left\{r_{i}^{\textrm{in}}\times\mathcal{W}_1\times\mathcal{W}_2+\ve{b}_1\times\mathcal{W}_2+\ve{b}_2\right\}}_{i=1}^n.\\
\end{aligned}
\end{equation}

Then $\ve{r}_{t}^{\textrm{total}}$ is obtained by AIIR-MIX, which is introduced in Fig.~\ref{fig:2}\textcolor{blue}{b} in more detail.

\section{Experiments}
In this Section, we evaluate our AIIR-MIX method on the StarCraft Multi-Agent Challenge (SMAC) environment~\cite{samvelyan2019starcraft}, which has become a benchmark for evaluating MARL methods. We compare AIIR-MIX with the state-of-the-art MARL methods such as LIIR~\cite{du2019liir}, QMIX~\cite{rashid2018qmix}, COMA~\cite{foerster2018counterfactual}, QTRAN~\cite{son2019qtran}. We conduct ablation experiments to demonstrate the effectiveness and rationality of AIIR and Mixing network. Finally, in order to analyze the learning process of the agents more clearly, we visualize the attention weights and intrinsic reward at each timestep.

\subsection{StarCraft II Micromanagement}
The StarCraft Multi-Agent Challenge (SMAC) environment is an experimental environment based on the real-time strategy game StarCraft II. Compared with the full StarCraft II, it focuses more on the micro-strategy of each agent than on macro-operations, that is, SMAC focuses on how to control each agent to defeat the enemy without considering the high-level macro-operations such as how to develop the economy and perform resource scheduling.
\begin{table}[ht]
\centering
\caption{Maps in different scenarios.}
\label{table-Maps}
\begin{tabular}{llll}
\toprule
Name & Ally Units                                                                & Enemy Units                                                               & Type~ ~        \\
\midrule
2s3z & 2 Stalkers 3 Zealots                                                      & 2 Stalkers 3 Zealots                                                      & heterogeneous  \\
3s5z & 3 Stalkers 5 Zealots                                                      & 3 Stalkers 5 Zealots                                                      & heterogeneous  \\
8m   & 8 Marines                                                                 & 8 Marines                                                                 & homogeneous    \\
MMM  & \begin{tabular}[c]{@{}l@{}}1 Medivac 2 Marauders\\7 Marines~\end{tabular} & \begin{tabular}[c]{@{}l@{}}1 Medivac 2 Marauders \\7 Marines\end{tabular} & heterogeneous  \\
\bottomrule
\end{tabular}
\end{table}

In the experiment, this paper applies all the default settings in SMAC, including game difficulty settings, shooting range and observation range. Both the shooting range and the observation range are circles with a certain radius. Only the agents within the observation range can enter the field of view, and only the agents within the shooting range can be attacked. As shown in the environment of Fig.~\ref{fig:1}, the attributes of the agents include weapon cooling down (CD), health point (HP), shield (2S3Z and 3S5Z), unit type, relative distance of the unit being observed, and last action. The action space of an agent consists of four discrete actions: move[direction], attack[enemy id], stop and noop. The agent movement space includes four directions: east, south, west and north. The attack action requires designating the enemy id within its shooting range. We select four challenging symmetric scenarios, 2s3z, 3s5z, MMM and 8m to evaluate the performance of the algorithms. 2s3z, 3s5z and MMM are heterogeneous maps and 8m is a homogeneous map. The scenario details for different maps of SMAC are shown in Table ~\ref{table-Maps}. Different game characters have different health point, attack power and shooting range.

We train these methods in 5 independent runs with different random seeds. During each run, these methods are evaluated every 5000 timesteps of training in 20 independent evaluation episodes. The evaluation episode will be regarded as the winning episode if all enemy units are defeated in time limit, then the percentage of the winning episodes is calculated as the win rate.

\begin{figure}[t]
\centering
\subfigure[8m]{\includegraphics[scale=0.365]{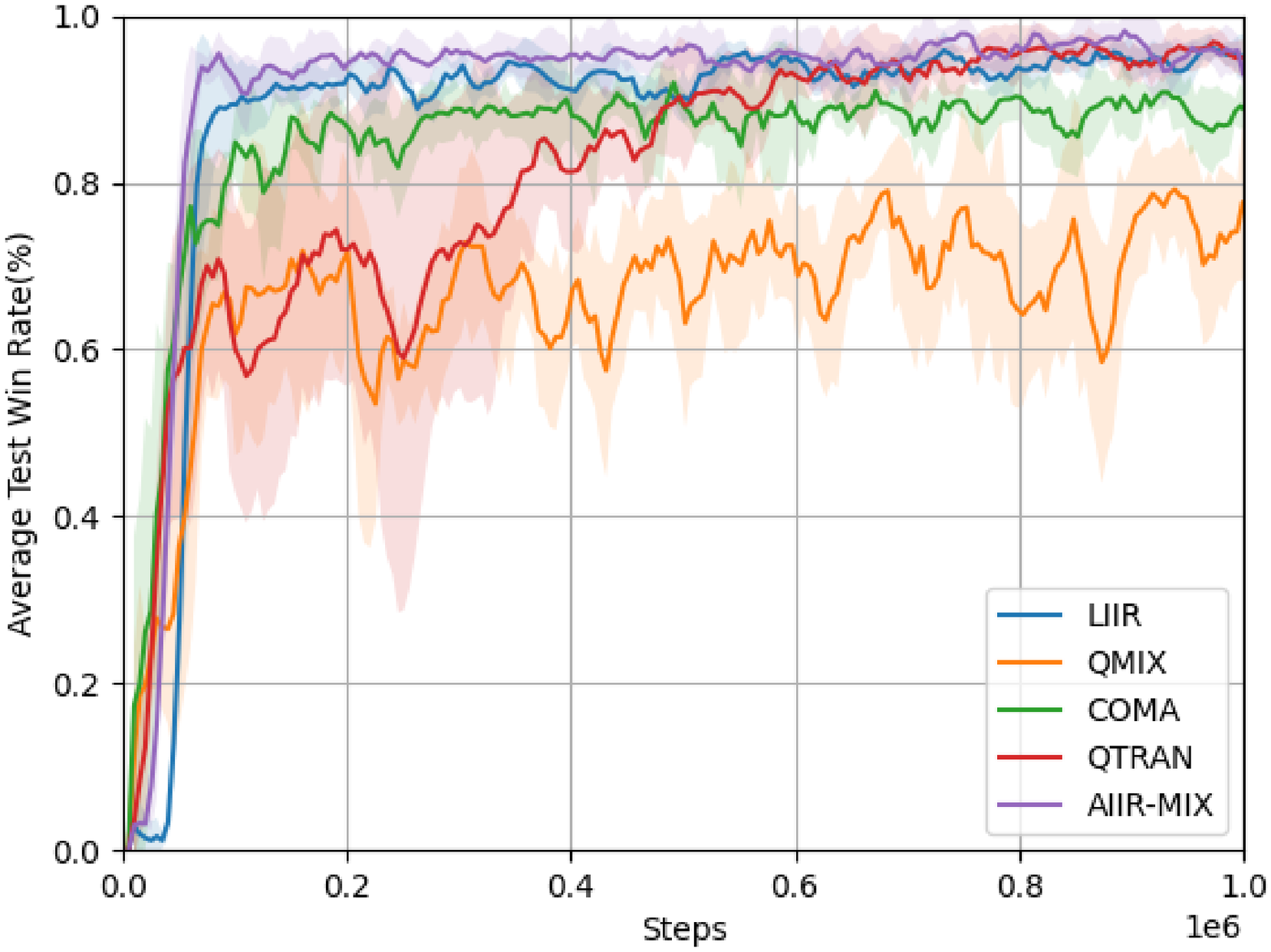}
}
\hspace{0mm}
\subfigure[MMM]{\includegraphics[scale=0.365]{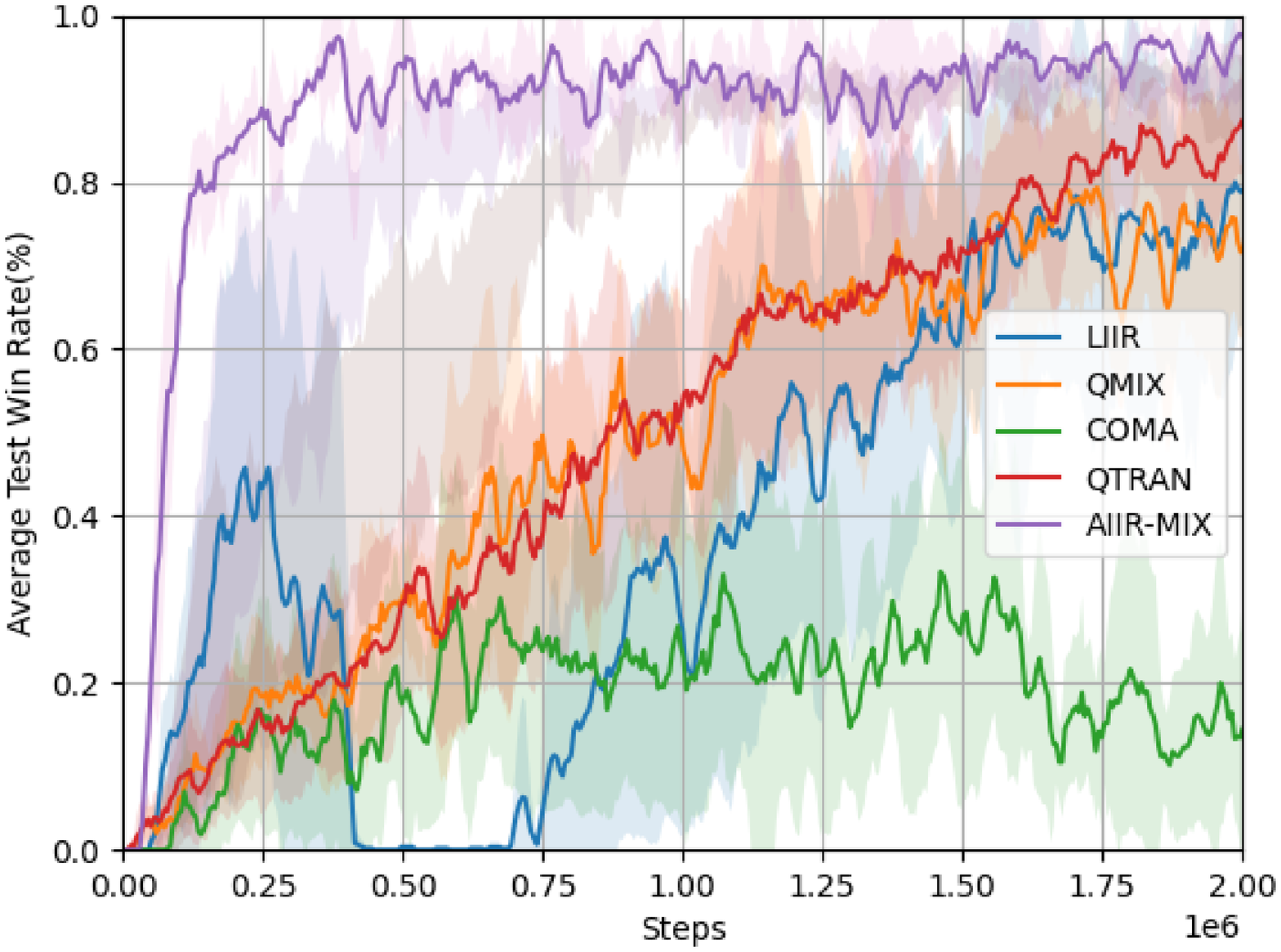}
}
\hspace{0mm}
\subfigure[2s3z]{\includegraphics[scale=0.365]{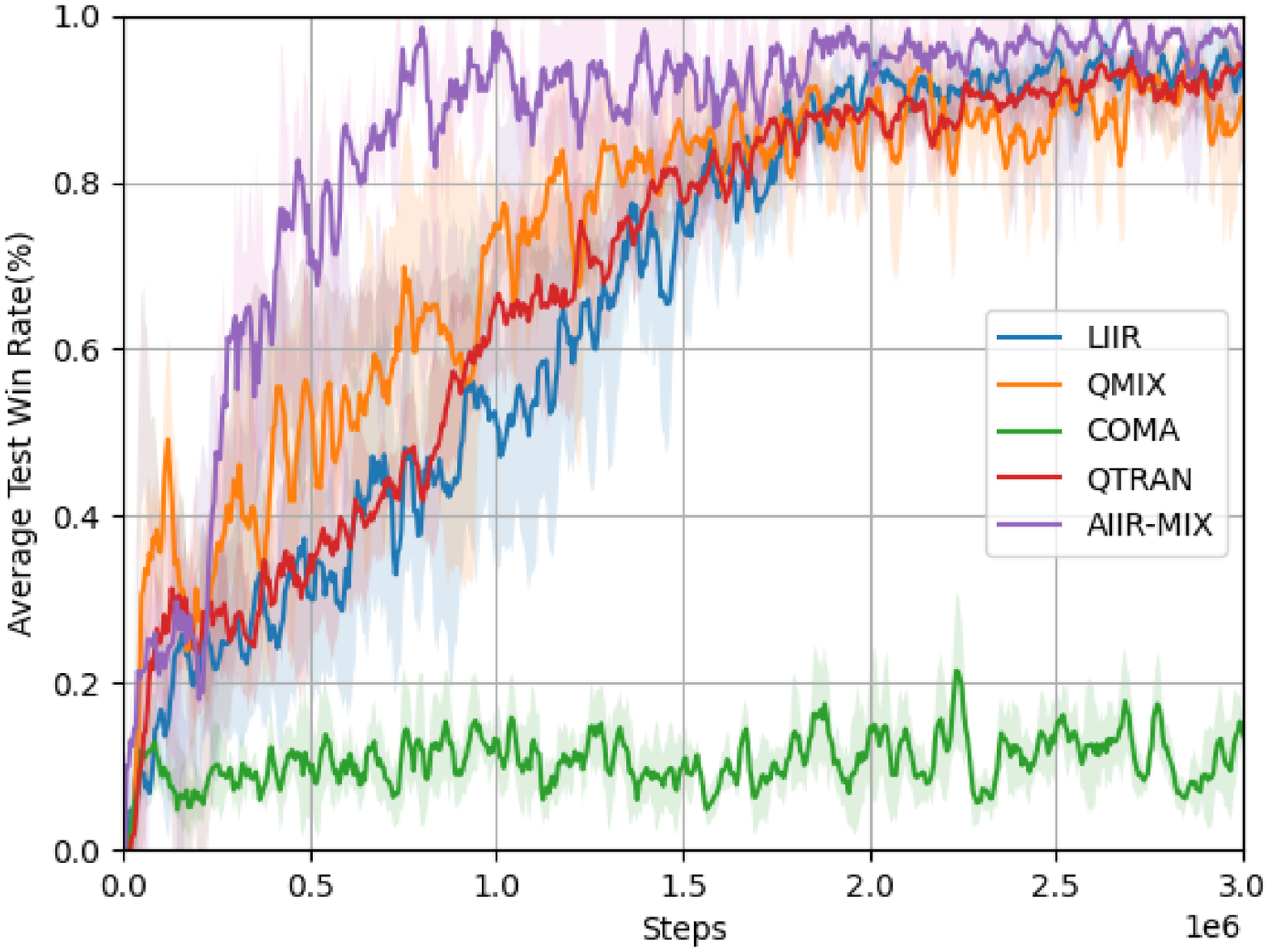}
}
\hspace{0mm}
\subfigure[3s5z]{\includegraphics[scale=0.365]{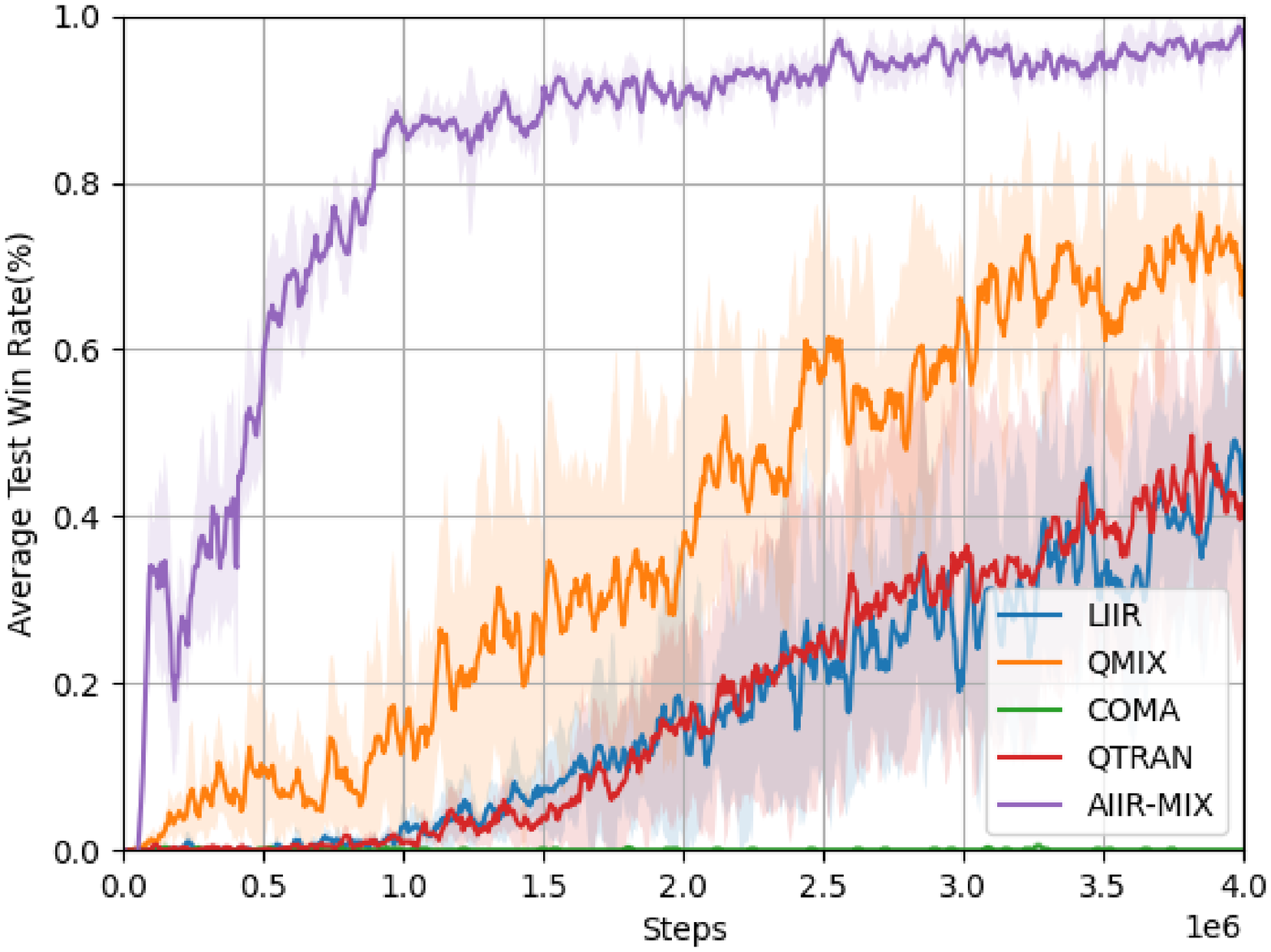}
}
\caption{Average Test Win Rate for LIIR, QMIX, COMA, QTRAN, and AIIT-MIX on Homogeneous and Heterogeneous Scenarios
\label{fig:3}}
\end{figure}
\subsection{Comparison \label{5.2}}
Fig.~\ref{fig:3} demonstrates the performance on 4 different maps in SMAC. Among all the baseline algorithms, QMIX and QTRAN are the most advanced algorithms among the current predominant value decomposition algorithms. In MMM and 2s3z, both QMIX and QTRAN obtain good performance. QMIX obtains a good performance in 3s5z. However, QMIX and QTRAN converge at a much slower pace than other state-of-the-art algorithms and QMIX performs relatively poorly in 8m. These results show that QMIX and QTRAN based on value decomposition can master the heterogeneous scenarios to some extent, but has a relatively poor performance in homogeneous scenarios. COMA fails to perform well in MMM, 2s3z, and 3s5z scenarios, and this result confirms that there is a performance gap between counterfactual policy gradients and TD dominance policy gradients in guiding the Actor network. LIIR can master homogeneous scenarios, such as 8m, but in heterogeneous scenarios, such as in MMM, 2s3z, and 3s5z, it has a slow convergence speed and poor performance. For all scenarios, the AIIR-MIX algorithm consistently outperforms the other algorithms. This result concludes that the Mixing network, which combines the intrinsic reward generated by the attention mechanism and extrinsic reward, can significantly contribute to obtaining better trained policies.

\subsection{Ablations}
\begin{figure}[t]
\centering
\subfigure[8m]{\includegraphics[scale=0.365]{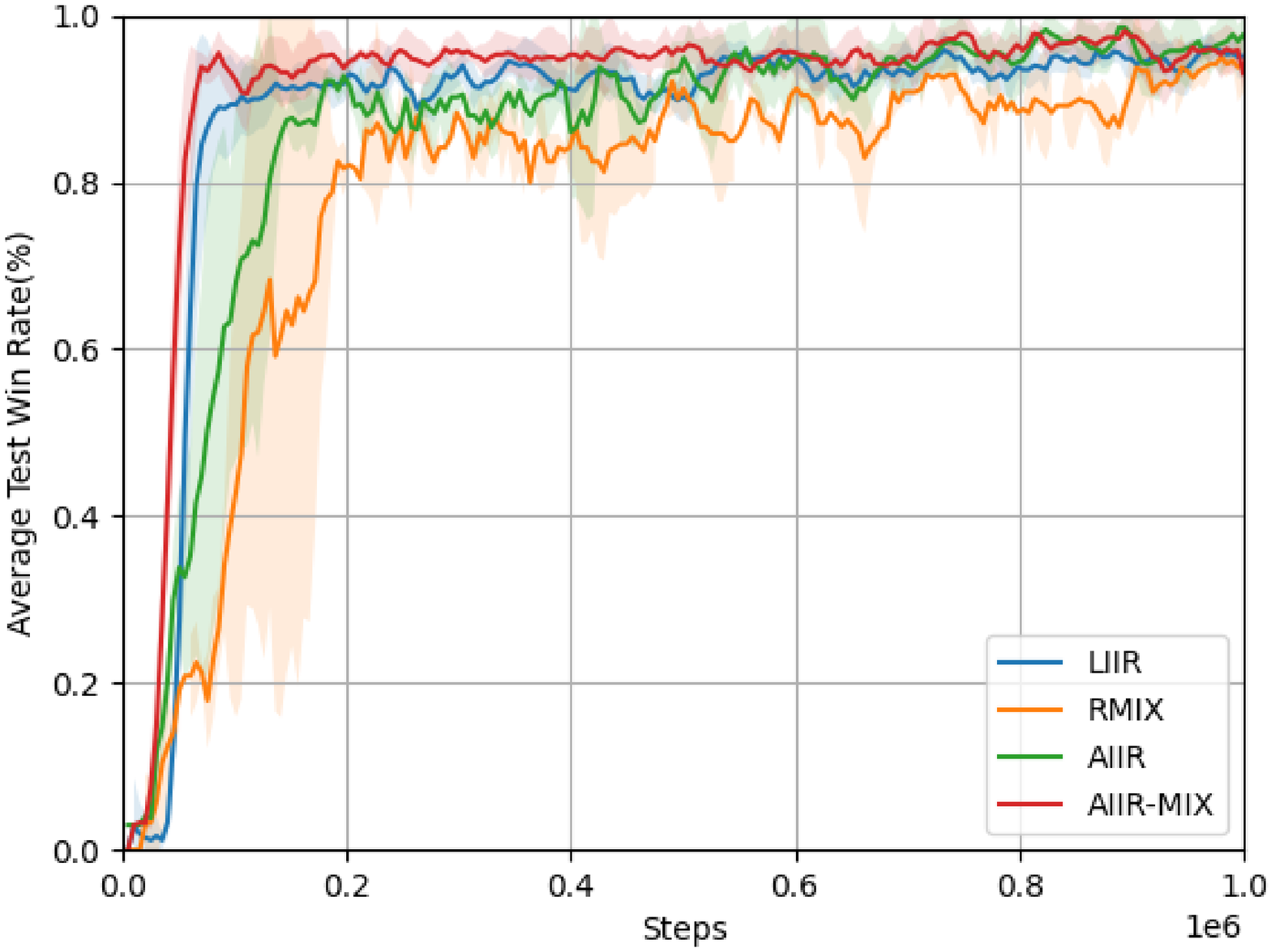}
}
\hspace{0mm}
\subfigure[2s3z]{\includegraphics[scale=0.365]{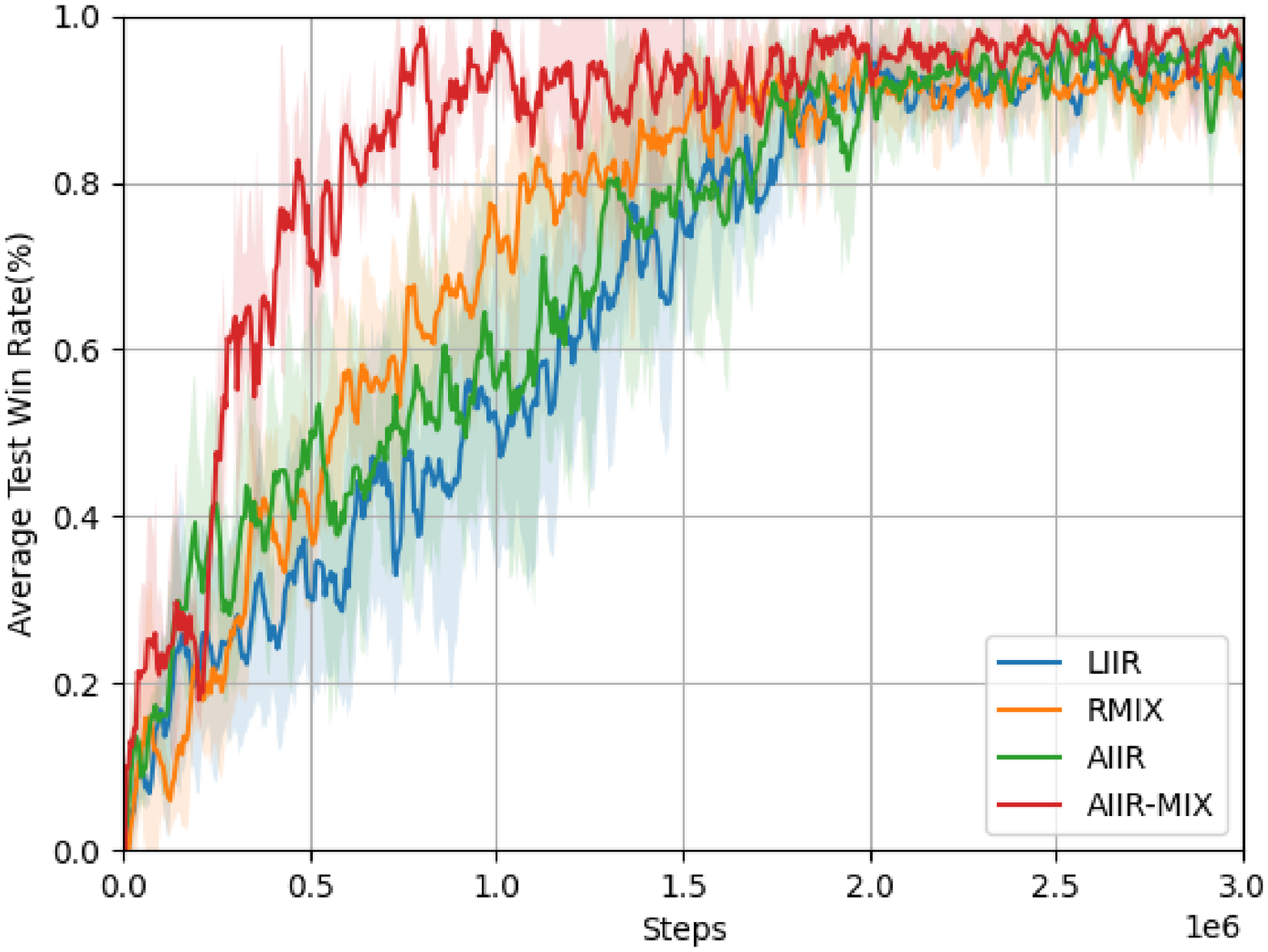}
}
\caption{Average Test Win Rate for AIIR-MIX and ablations on 8m and 2s3z.
\label{fig:4}}
\end{figure}
We conduct ablation experiments to investigate the effect of the attention mechanism on AIIR and the necessity of non-linear transformation in Mixing network. For one thing, we analyze the importance of the attention mechanism in AIIR by comparing it with RMIX (i.e., Mixing network), which employs the same intrinsic reward network structure as in LIIR instead of an attention mechanism to generate intrinsic reward. For another, we investigate the necessity of the non-linear Mixing network. We replace the Mixing network with a linear Mixing network in which intrinsic and extrinsic reward of each agent are simply weighted and summed to generate the global reward.

\begin{figure}[t]
\centering
\subfigure[Intrinsic reward]{\includegraphics[scale=0.49,trim={0cm 0cm 0cm 0cm},clip]{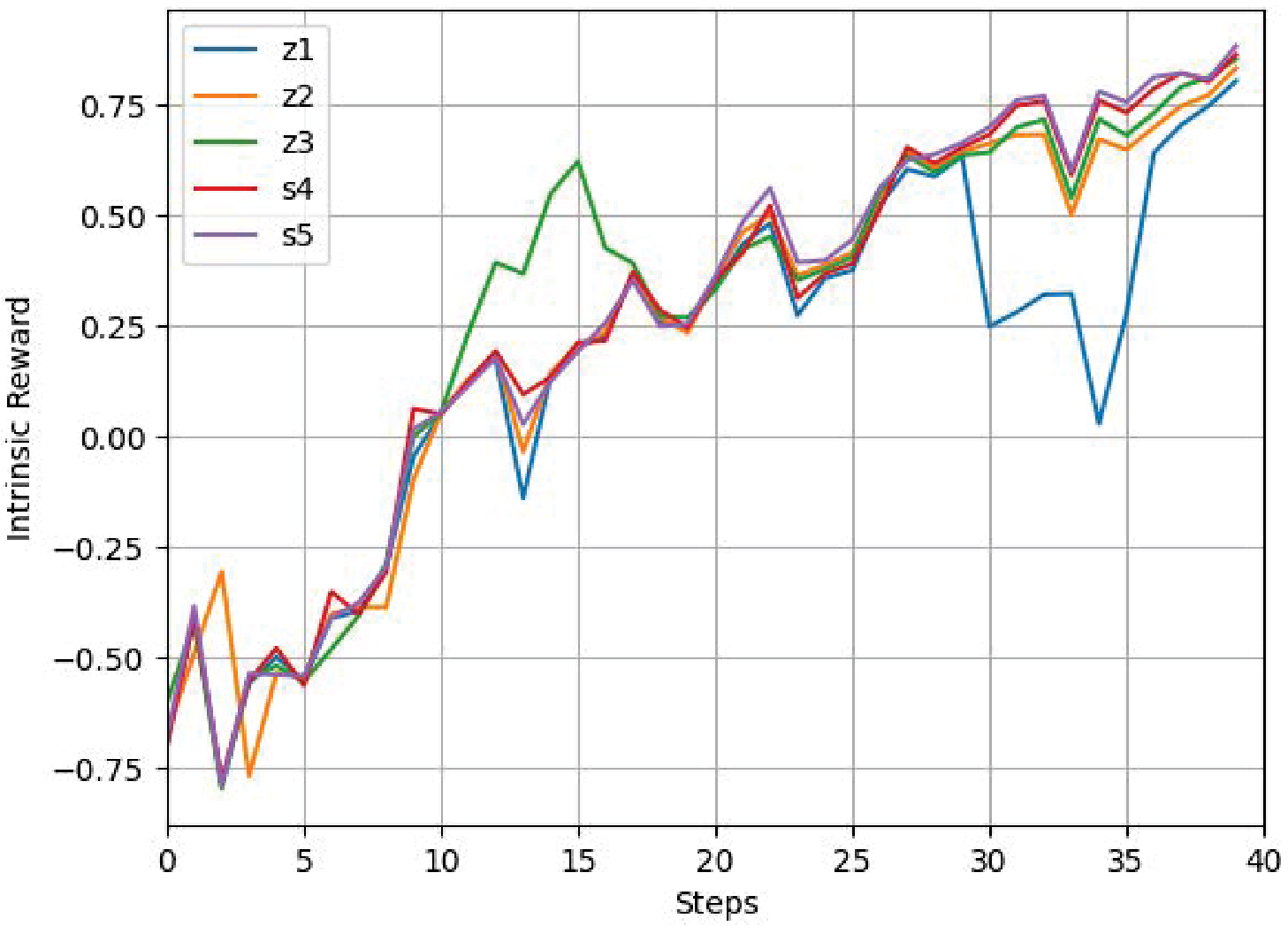}
}
\hspace{17pt}
\subfigure[Attention weights]{\includegraphics[width=0.4\textwidth]{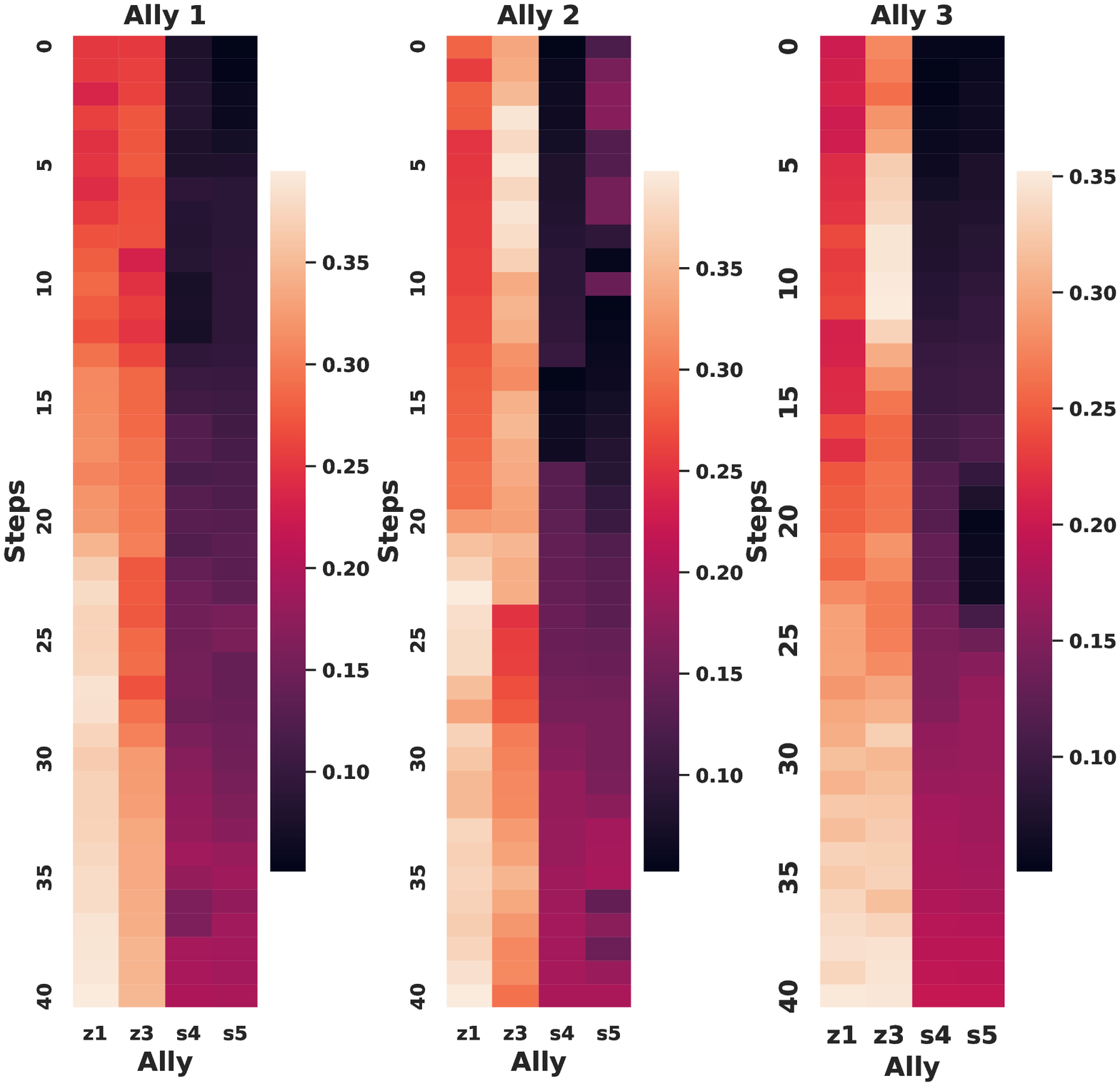}\hspace{21pt}}
\caption{Intrinsic Reward and Attention Weights in AIIR-MIX.
\label{fig:5}}
\end{figure}

Fig.~\ref{fig:4} shows the results of AIIR-MIX and its ablations on SMAC benchmark 8m and 2s3z. It can be seen that AIIR-MIX outperforms all its ablations in the experiments. Fig.~\ref{fig:4}\textcolor{blue}{a} shows that in homogeneous scenarios, RMIX and AIIR slow down the learning speed compared with the baselines algorithm LIIR due to the complexity of the architecture. However, when the attention-based intrinsic reward function and the non-linear Mixing network are together in AIIR-MIX, the improvement is capable of making up for the decrease in training speed results from the complex architecture. Fig.~\ref{fig:4}\textcolor{blue}{b} shows that in heterogeneous scenarios, a non-linear Mixing network is required to achieve better performance. Also, the performance of AIIR-MIX compared with RMIX demonstrates the importance of precise intrinsic reward and the combination with extrinsic reward through a non-linear approach.

\subsection{Visualization}
\begin{figure}[t]
\centering
\subfigure[t=5]{\includegraphics[scale=0.45]{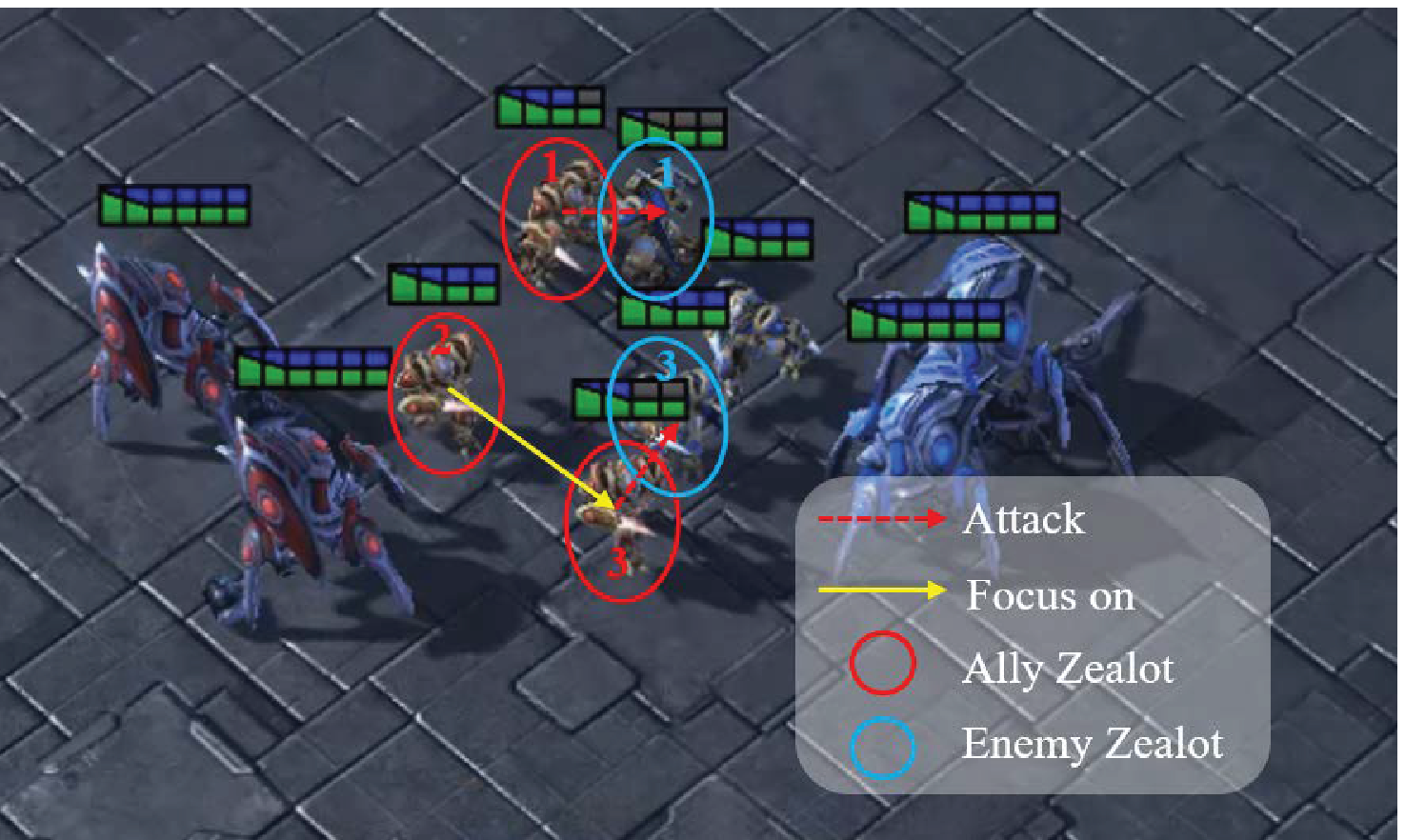}
}
\hspace{0mm}
\subfigure[t=15]{\includegraphics[scale=0.45]{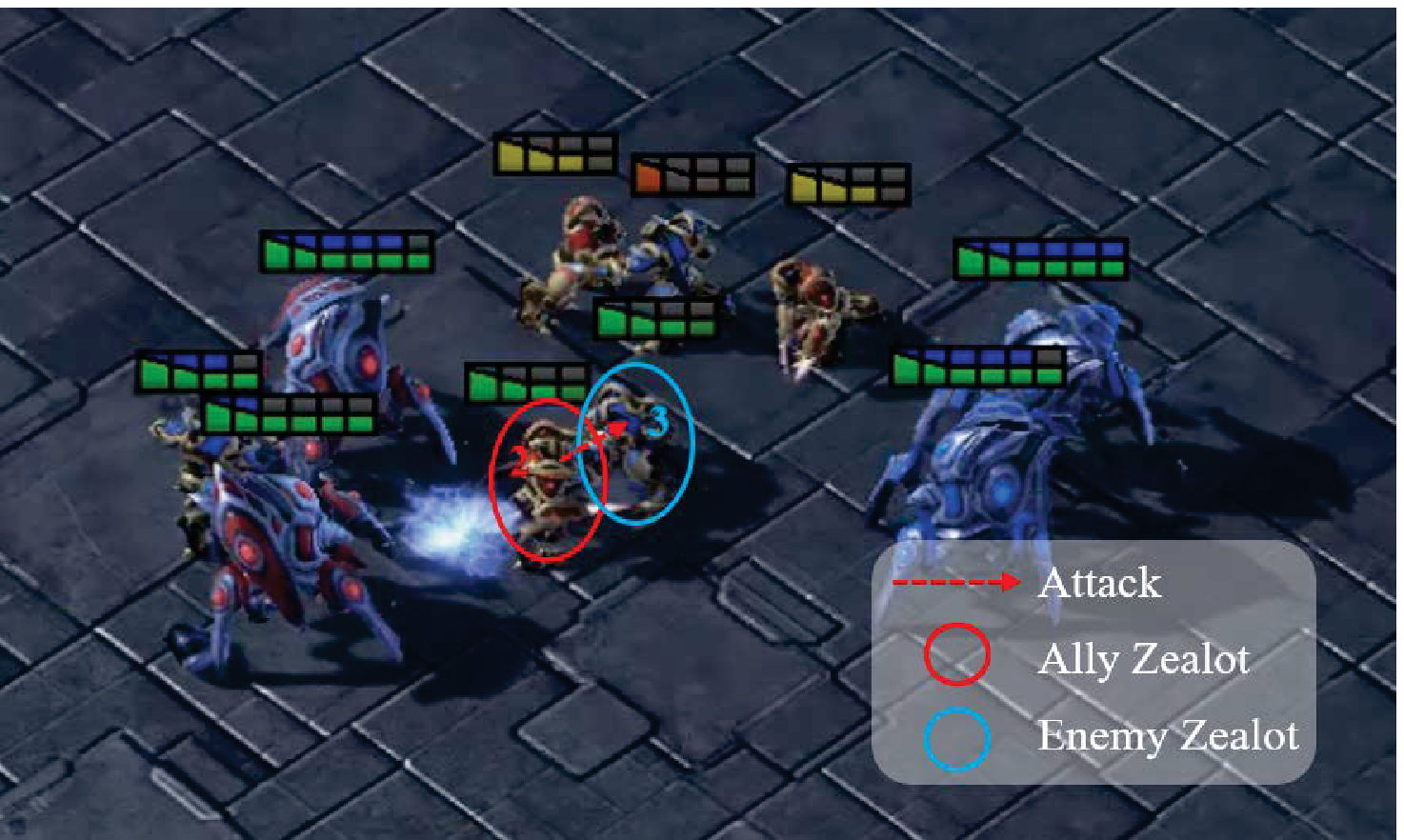}
}
\hspace{0mm}
\subfigure[t=23]{\includegraphics[scale=0.45]{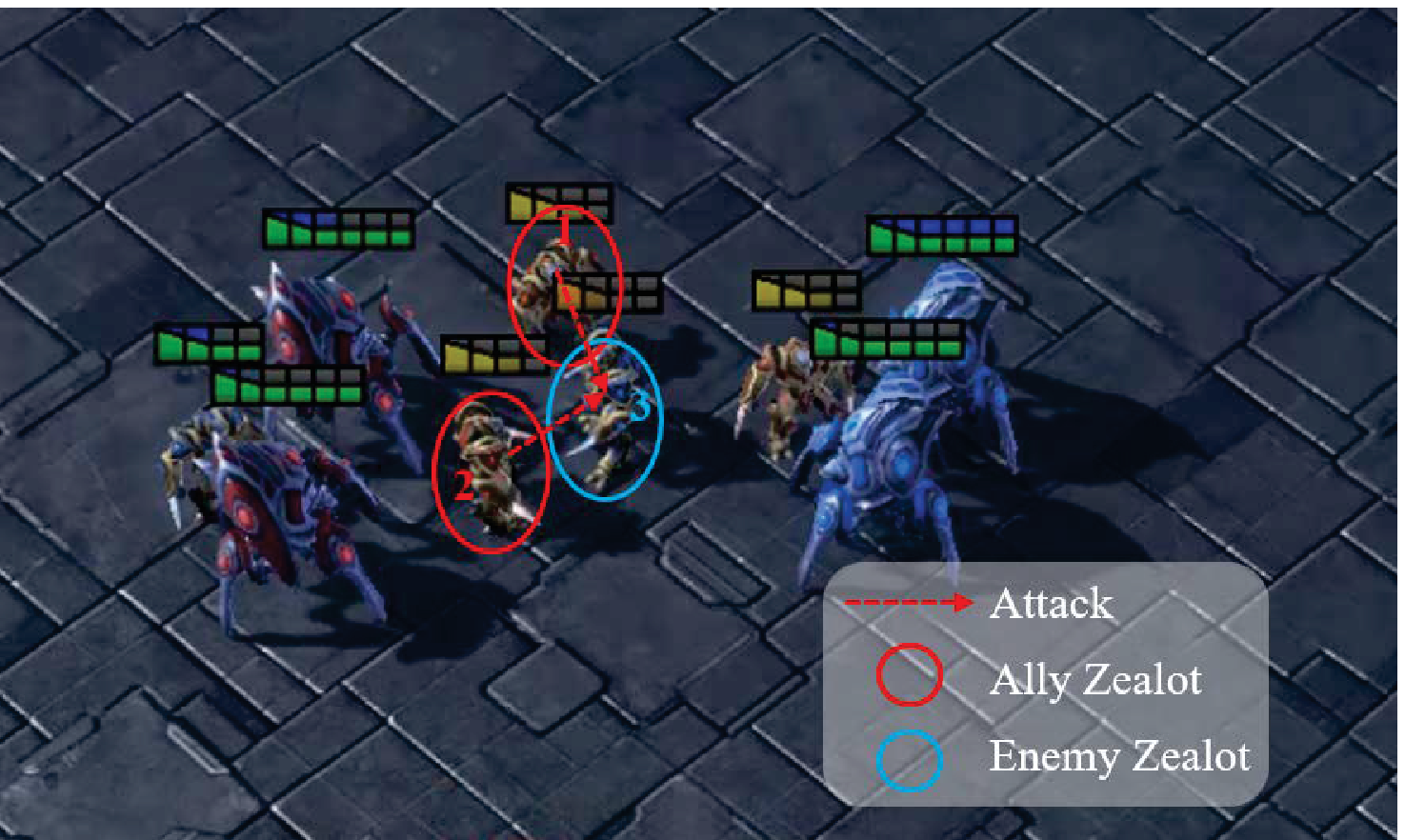}
}
\hspace{0mm}
\subfigure[t=34]{\includegraphics[scale=0.45]{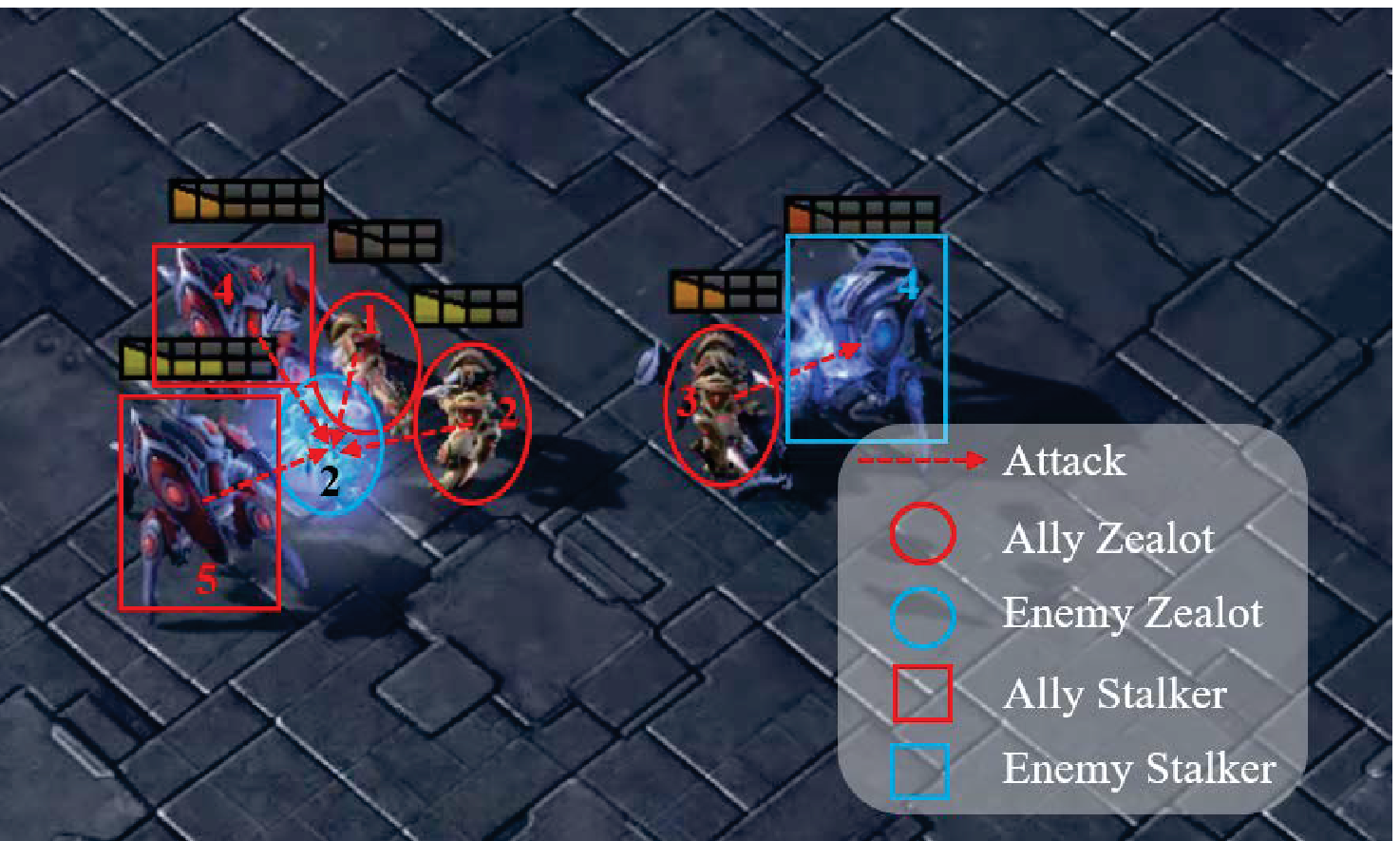}
}
\caption{Some auxiliary snapshots in 2s3z.
\label{fig:6}}
\end{figure}

In addition to evaluate the performance of the trained policy in Section \ref{5.2}, we are more curious about how much the attention mechanism contributes to learning intrinsic reward, and the learned intrinsic reward contributes to policy learning. In order to figure out what is learned in the above two processes, we propose to explicitly visualize the attention weights and intrinsic reward at each timestep in the trajectory. For clarity, we choose scenario 2s3z which has a relatively small number of agents for our analysis. Fig.~\ref{fig:5} shows the attention weights and intrinsic rewards of the agents. And Fig.~\ref{fig:6} shows some auxiliary snapshots of the agents. The above figures include agent type (s means Stalker, z means Zealot) and agent id (from 1 to 5).

It can be seen in Fig.~\ref{fig:5}\textcolor{blue}{a} that after timestep 10, the intrinsic reward of ally Zealot 3 increases a lot. The reason for this phenomenon is clearly illustrated in Fig.~\ref{fig:6}\textcolor{blue}{a} and Fig.~\ref{fig:6}\textcolor{blue}{b}. Before timestep 10, ally Zealot 3 with a low HP is attacking the enemy Zealot 3. After timestep 10, while ally Zealot 2 starts attacking the enemy, ally Zealot 3 stops firing and flees. Avoiding attacking the enemy head-on is certainly a good behavior when the agent does not have enough HP. After timestep 30, the intrinsic reward of ally Zealot 1 decreases a lot. As shown in Fig.~\ref{fig:6}\textcolor{blue}{d}, ally Zealot 1 is attacking enemy Zealot 2 along with ally Zealot 2, 4 and 5. However, ally Zealot 1's HP is very low at this point, and it is not a good policy to continue attacking the enemy. In general, the intrinsic reward increases when the agent chooses a good policy, and decreases when the agent chooses a bad policy. This more precise reward contributes significantly in helping the agent learn good policies quickly.

As shown in Fig.~\ref{fig:5}\textcolor{blue}{a}, the intrinsic reward are more fluctuating for Zealot-type agents. For a more detailed analysis, we visualize the attention weights for Zealot-type agents. Fig.~\ref{fig:5}\textcolor{blue}{b} shows the attention weights of three Zealots to other allied agents. It is clear that Zealots always pay more attention to the other two Ally Zealots. The elevated attention weights indicate the occurrence of cooperative behavior between agents, including ally Zealot 1 to ally Zealot 2 at timestep 23 and ally Zealot 2 to ally Zealot 3 at timestep 5 in Fig.~\ref{fig:6}. That is, the attention mechanism facilitates the cooperative behavior between agents in a certain extent.

\section{Conclusion}
This paper presents AIIR-MIX, a novel multi-agent RL algorithm that learns an individual intrinsic reward through an attention mechanism for each agent so that the agent can obtain different rewards to facilitate the learning of the agent. Besides, we design a non-linear Mixing network to combine intrinsic and extrinsic rewards instead of a simple linear summation for the first time, thereby dynamically generating global rewards for each agent according to the changing environment. Our empirical results of the experiments carried out on the battle games in StarCraft II demonstrate that our approach induces trained policy compared with a few state-of-the-art MARL methods better. We further conduct ablation studies to confirm the effectiveness of the intrinsic reward network based on the attention mechanism and the non-linear Mixing network. And we visualize the intrinsic rewards and attention weights to illustrate how the intrinsic reward network assigns each agent an appropriate reward and promotes cooperation amongst agents.

\acks{This work was supported in part by the Aeronautical Science Foundation of China under Grant 20200058069001 and in part by the Fundamental Research Funds for the Central Universities under Grant 2242021R41094.}

\bibliography{reference}
\end{sloppypar}
\end{document}